\documentstyle[colap,psfig]{article}    
\title{Three New Probabilistic Models \\ for Dependency Parsing: An Exploration%
\thanks{This material is based upon work supported under a National
Science Foundation Graduate Fellowship, and has benefited greatly from
discussions with Mike Collins, Dan Melamed, Mitch Marcus and Adwait Ratnaparkhi.}}
\author{Jason M. Eisner \\
CIS Department, University of Pennsylvania \\
200 S. 33rd St., Philadelphia, PA 19104-6389, USA \\
{\tt jeisner@linc.cis.upenn.edu}}

\newcommand\longcomment[1]{}
\newcommand\refp[1]{(\ref{#1})}

\begin{document}
\maketitle
\begin{abstract}
\small After presenting a novel $O(n^3)$ parsing algorithm for
dependency grammar, we develop three contrasting ways to stochasticize
it.  We propose (a) a lexical affinity model where words struggle to
modify each other, (b) a sense tagging model where words fluctuate
randomly in their selectional preferences, and (c) a generative model
where the speaker fleshes out each word's syntactic and conceptual
structure without regard to the implications for the hearer.  We also
give preliminary empirical results from evaluating the three models'
parsing performance on annotated {\em Wall Street Journal} training
text (derived from the Penn Treebank).  In these results, the
generative model performs significantly better than the others, and
does about equally well at assigning part-of-speech tags.
\end{abstract}

\bibliographystyle{acl}

\section{Introduction} 

In recent years, the statistical parsing community has begun to reach
out for syntactic formalisms that recognize the individuality of
words.  Link grammars \cite{Sleator+Temperley} and lexicalized tree-adjoining
grammars \cite{Schabes} have now received stochastic treatments.  Other
researchers, not wishing to abandon context-free grammar (CFG) but
disillusioned with its lexical blind spot, have tried to re-parameterize
stochastic CFG in context-sensitive ways \cite{Black+al} or have augmented the
formalism with lexical headwords \cite{Magerman,Collins}.

\begin{figure}
\centerline{\psfig{figure=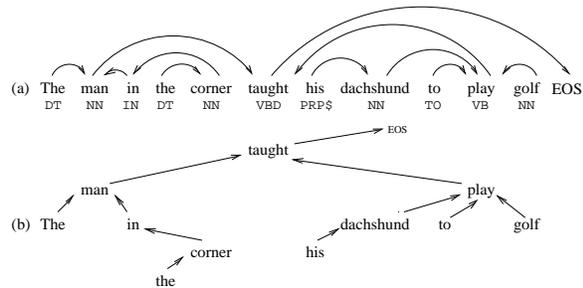,width=3in}}
\caption{\small (a) A bare-bones dependency parse.  Each word points to a single
{\bf parent}, the word it modifies; the head of the sentence
points to the {\bf EOS} (end-of-sentence) mark.  Crossing links and cycles are not allowed.
(b) Constituent structure
and subcategorization may be highlighted by displaying the same dependencies as a lexical tree.}\label{sampleparse}
\end{figure}

In this paper, we present a flexible probabilistic parser that
simultaneously assigns both part-of-speech tags and a bare-bones dependency structure
(illustrated in Figure~\ref{sampleparse}%
).  The
choice of a simple syntactic structure is deliberate: we would like to
ask some basic questions about where lexical relationships appear and how best to 
exploit them.  It is useful to look into these basic questions before trying
to fine-tune the performance of systems whose behavior is harder to
understand.%
\footnote{Our novel parsing algorithm also rescues dependency from
certain criticisms: ``Dependency grammars \ldots are not lexical, and (as far as we
know) lack a parsing algorithm of efficiency comparable to link
grammars.'' \cite[p.~3]{Lafferty+al}
}

The main contribution of the work is to propose three distinct,
lexicalist hypotheses about the probability space underlying
sentence structure.  We illustrate how each hypothesis is expressed
in a dependency framework, and how each can be used to guide our
parser toward its favored solution.  Finally, we point to experimental
results that compare the three hypotheses' parsing performance on
sentences from the {\em Wall Street Journal}.  The parser is trained
on an annotated corpus; no hand-written grammar is required.

\section{Probabilistic Dependencies}


It cannot be emphasized too strongly that a {\em grammatical
representation} (dependency parses, tag sequences, phrase-structure
trees) does not entail any particular {\em probability model}.  In
principle, one could model the distribution of dependency parses in
any number of sensible or perverse ways.  The choice of the right
model is not {\em a priori} obvious.

One way to build a probabilistic grammar is to specify what sequences
of moves (such as shift and reduce) a parser is likely to make.  It is
reasonable to expect a given move to be correct about as often on test
data as on training data.  This is the philosophy behind stochastic
CFG \cite{Jelinek+al}, ``history-based'' phrase-structure parsing
\cite{Black+al}, and others.  

However, probability models derived from parsers sometimes focus on
incidental properties of the data.
This may be the case for \cite{Lafferty+al}'s model for link grammar.
If we were to adapt their top-down stochastic parsing strategy to the
rather similar case of dependency grammar, we would find their
elementary probabilities tabulating only non-intuitive aspects of the
parse structure:
\begin{center}
$Pr($word $j$ is the rightmost pre-$k$
child of word $i$ $\mid$ $i$ is a right-spine strict descendant of one
of the left children of a token of word $k$, or else $i$ is the parent
of $k$, and $i$ precedes $j$ precedes $k$).\footnote{This corresponds
to Lafferty et al.'s central statistic (p.\ 4), $\Pr(W, \leftarrow \mid L,R,l,r)$, in
the case where $i$'s parent is to the left of $i$.  $i, j, k$ correspond
to $L, W, R$ respectively.  Owing to the particular recursive strategy the parser uses to break up
the sentence, the statistic would be measured and utilized only under the condition described above.}
\end{center}
\noindent While it is clearly necessary to decide whether $j$ {\em is}
a child of $i$, conditioning that decision as above may not reduce its
test entropy as much as a more linguistically perspicuous condition
would.


We believe it is fruitful to design probability models
independently of the parser.  In this section, we will outline the
three lexicalist, linguistically perspicuous, qualitatively different
models that we have developed and tested.

\subsection{Model A: Bigram lexical affinities}\label{sec:indeplink}

$N$-gram taggers like  
\cite{Church:trigram,Jelinek:trigram,Kupiec:trigram,Merialdo:trigram}
take the following view of how a tagged sentence enters
the world.  First, a sequence of tags is generated according to a
Markov process, with the random choice of each tag conditioned on the
previous two tags.  Second, a word is chosen conditional on each tag.  

\begin{figure*}
\begin{eqnarray}
\lefteqn{Pr(\underline{\rm words, tags, links}) = Pr({\rm words, tags}) 
                                       \cdot Pr({\rm link\ presences\ and\ absences} 
                                               \mid {\rm words, tags})} 
                                            \label{indepmodelparseprob} \\
    & \approx & \prod_{1\leq i\leq n} Pr({\sl tword}(i) \mid {\sl tword}(i+1), {\sl tword}(i+2)) 
                       \cdot \prod_{1\leq i,j \leq n} Pr(L_{ij} \mid {\sl tword}(i), {\sl tword}(j))~~~\hspace*{0.7in}\ 
                           \label{tagexpansion} \\
\lefteqn{Pr({\sl tword}(i)  \mid {\sl tword}(i+1), {\sl tword}(i+2)) 
  \approx  Pr({\sl tag}(i)  \mid {\sl tag}(i+1), {\sl tag}(i+2)) \cdot Pr({\sl word}(i) \mid {\sl tag}(i))}  \label{tagexpansionapprox} \\  
\lefteqn{Pr(\underline{\rm words, tags, links}) \propto Pr({\rm words, tags, preferences}) =  Pr({\rm words, tags}) \cdot Pr({\rm preferences \mid words, tags})} \label{selectmodelparseprob} \\
& \approx & \prod_{1\leq i\leq n} Pr({\sl tword}(i) \mid {\sl tword}(i+1), {\sl tword}(i+2)) \cdot \prod_{1\leq i\leq n} Pr({\sl preferences}(i) \mid {\sl tword}(i)) \nonumber \\
\lefteqn{Pr(\underline{\rm words, tags, links})  = \prod_{1\leq i\leq n} \left(\prod_{c=-(1+{\sl \scriptsize \#left\mbox{\sl -}kids}(i)), c \neq 0}^{1+{\sl \scriptsize \#right\mbox{\sl -}kids}(i)} \hspace{-0.5in} Pr({\sl tword}({\sl kid}_c(i)) \mid {\sl tag}(\begin{array}[t]{c}{\sl kid}_{c - 1}(i) \\ \makebox[0in]{\scriptsize or ${\sl kid}_{c+1}$ if $c<0$} \end{array}), {\sl tword}(i) \right)} \label{genmodelparseprob} 
\end{eqnarray} 
\caption{\small High-level views of model A (formulas \ref{indepmodelparseprob}--\ref{tagexpansionapprox}); model B (formula \ref{selectmodelparseprob}); and model C (formula \ref{genmodelparseprob}).  If $i$ and $j$ are tokens,
then ${\sl tword}(i)$ represents the pair $({\sl tag}(i), {\sl word}(i))$, and $L_{ij} \in \{0,1\}$ is 1 iff $i$ is the 
parent of $j$.}
\label{formulas}
\end{figure*} 

Since our sentences have links as well as tags and words, suppose that
after the words are inserted, each sentence passes through a third
step that looks at each pair of words and randomly decides whether to
link them.  For the resulting sentences to resemble real corpora, the
probability that word $j$ gets linked to word $i$ should be {\em
lexically sensitive}: it should depend on the (tag,word) pairs at both
$i$ and $j$.

The probability of drawing a given parsed sentence from the population
may then be expressed as \refp{indepmodelparseprob} in
Figure~\ref{formulas}, where the random variable $L_{ij} \in \{0,1\}$
is 1 iff word $i$ is the parent of word $j$.

Expression \refp{indepmodelparseprob} assigns a probability to every
possible tag-and-link-annotated string, and these probabilities sum to
one.  Many of the annotated strings exhibit violations such as
crossing links and multiple parents---which, if they were allowed,
would let all the words express their lexical preferences
independently and simultaneously.  We stipulate that the model
discards from the population any illegal structures that it generates;
they do not appear in either training or test data.  Therefore, the
parser described below finds the likeliest {\em legal} structure: it
maximizes the lexical {\em preferences} of \refp{indepmodelparseprob}
within the few hard linguistic {\em constraints} imposed by the
dependency formalism.

\longcomment{!!! clarify? move? cut?}  In practice, some
generalization or ``coarsening'' of the conditional probabilities in
\refp{indepmodelparseprob} helps to avoid the effects of
undertraining.  For example, we follow standard practice
\cite{Church:trigram} in $n$-gram tagging by using
\refp{tagexpansionapprox} to approximate the first term in
\refp{tagexpansion}.  Decisions about how much coarsening to do are of
great practical interest, but they depend on the training corpus and
may be omitted from a conceptual discussion of the model.

\begin{figure}
\centerline{\psfig{figure=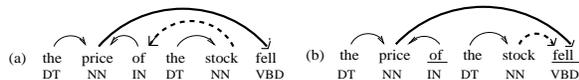,width=3in}}
\caption{\small (a) The correct parse.  (b) A common error if the model ignores arity.}\label{stock}
\end{figure}

The model in \refp{indepmodelparseprob} can be improved; it does not
capture the fact that words have arities.  For example, {\em the price
of the stock fell} (Figure~\ref{stock}a) will typically be misanalyzed
under this model.  Since stocks often fall, {\em stock} has a greater
affinity for {\em fell} than for {\em of}.  Hence {\em stock} (as well
as {\em price}) will end up pointing to the verb {\em fell}
(Figure~\ref{stock}b), resulting in a double subject for {\em fell}
and leaving {\em of} childless.  To capture word arities and other
subcategorization facts, we must recognize that the children of a word
like {\em fell} are not independent of each other.

The solution is to modify \refp{indepmodelparseprob} slightly, further
conditioning $L_{ij}$ on the number and/or type of children of $i$
that already sit between $i$ and $j$.  This means that in the parse of
Figure~\ref{stock}b, the link {\em price} $\rightarrow$ {\em fell}
will be sensitive to the fact that {\em fell} already has a closer
child tagged as a noun (NN).  Specifically, the {\em price}
$\rightarrow$ {\em fell} link will now be strongly disfavored in
Figure~\ref{stock}b, since verbs rarely take two NN dependents to the
left.  By contrast, {\em price} $\rightarrow$ {\em fell} is
unobjectionable in Figure~\ref{stock}a, rendering that parse more
probable.  (This change can be reflected in the conceptual model, by
stating that the $L_{ij}$ decisions are made in increasing order
of link length $|i-j|$ and are no longer independent.)


\subsection{Model B: Selectional preferences}

In a legal dependency parse, every word except for the head of the
sentence (the EOS mark) has exactly one parent.  Rather than having
the model select a subset of the $n^2$ possible links, as in model A,
and then discard the result unless each word has exactly one parent,
we might restrict the model to picking out one parent per word to
begin with.  Model B generates a sequence of tagged words, then
specifies a parent---or more precisely, a {\em type} of parent---for
each word $j$.

Of course model A also ends up selecting a parent for each word, but
its calculation plays careful politics with the set of other words
that happen to appear in the sentence: word $j$ considers both the
benefit of selecting $i$ as a parent, {\em and} the costs of spurning
all the other possible parents $i'$.%
 Model B takes an approach at the opposite extreme, and simply has
each word blindly describe its ideal parent.  For example, {\em price}
in Figure~\ref{stock} might insist (with some probability) that it
``depend on a verb to my right.''  To capture arity, words
probabilistically specify their ideal children as well: {\em fell} is
highly likely to want only one noun to its left.  The form and
coarseness of such specifications is a parameter of the model.  

When a word stochastically chooses one set of requirements on its
parents and children, it is choosing what a link grammarian would call
a {\em disjunct} (set of selectional preferences) for the word.  We
may thus imagine generating a Markov sequence of tagged words as
before, and then independently ``sense tagging'' each word with a
disjunct.\footnote{In our implementation, the distribution over
possible disjuncts is given by a pair of Markov processes, as in model
C.}  Choosing all the disjuncts does not quite specify a parse.
However, if the disjuncts are sufficiently specific, it specifies at
most one parse.  Some sentences generated in this way are illegal
because their disjuncts cannot be simultaneously satisfied; as in
model A, these sentences are said to be removed from the population,
and the probabilities renormalized.  A likely parse is therefore one
that allows a likely and consistent set of sense tags; its
probability in the population is given in \refp{selectmodelparseprob}.



\subsection{Model C:  Recursive generation}

The final model we propose is a {\bf generation} model, as opposed to
the {\bf comprehension} models A and B (and to other comprehension
models such as \cite{Lafferty+al,Magerman,Collins}).  The contrast
recalls an old debate over spoken language, as to whether its
properties are driven by hearers' acoustic needs (comprehension) or
speakers' articulatory needs (generation).  Models A and B suggest
that speakers produce text in such a way that the grammatical
relations can be easily decoded by a listener, given words'
preferences to associate with each other and tags' preferences to
follow each other.  But model C says that speakers' primary goal is to
flesh out the syntactic and conceptual structure for each word they
utter, surrounding it with arguments, modifiers, and function words as
appropriate.  According to model C, speakers should not hesitate to
add extra prepositional phrases to a noun, even if this
lengthens some links that are ordinarily short, or leads
to tagging or attachment ambiguities.

The generation process is straightforward.  Each time a word $i$ is
added, it generates a Markov sequence of (tag,word) pairs to serve as
its left children, and an separate sequence of (tag,word) pairs as its
right children.  Each Markov process, whose probabilities depend
on the word $i$ and its tag, begins in a special START
state; the symbols it generates are added as $i$'s children,
from closest to farthest, until it reaches the STOP state.  The
process recurses for each child so generated.  This is a sort of
lexicalized context-free model.

Suppose that the Markov process, when generating a child, remembers
just the tag of the child's most recently generated sister, if any.
Then the probability of drawing a given parse from the population is
\refp{genmodelparseprob}, where ${\sl kid}(i,c)$ denotes the
$c$th-closest right child of word $i$, and where ${\sl
kid}(i,0)=\mbox{START}$ and ${\sl kid}(i,1+\#right\mbox{\sl -}kids(i))
= \mbox{STOP}$.  ($c < 0$ indexes left children.)  This may be thought
of as a non-linear trigram model, where each tagged word is generated
based on the parent tagged word and a sister tag.  The links in the parse
serve to pick out the relevant trigrams, and are chosen to get
trigrams that optimize the global tagging.  That the links also happen
to annotate useful semantic relations is, from this perspective, quite
accidental.

Note that the revised version of model A uses probabilities $Pr($link
to child $\mid$ child, parent, closer-children), where model C uses
$Pr($link to child $\mid$ parent, closer-children).  This is because
model A assumes that the child was previously
generated by a linear process, and all that is necessary is to link to
it.  Model C actually generates the child in the process of linking to
it.

\section{Bottom-Up Dependency Parsing}

In this section we sketch our dependency parsing algorithm: a novel
dynamic-programming method to assemble the most probable parse from
the bottom up.
The algorithm adds one link at a time, making it easy to multiply out
the models' probability factors.  It also enforces the special
directionality requirements of dependency grammar, the prohibitions on
cycles and multiple parents.\footnote{Labeled dependencies are
possible, and a minor variant handles the simpler case of link
grammar.  Indeed, abstractly, the algorithm resembles a cleaner,
bottom-up version of the top-down link grammar parser developed
independently by \cite{Lafferty+al}.}

The method used is similar to the CKY method of context-free parsing, 
which combines analyses of shorter substrings into analyses of
progressively longer ones.  Multiple analyses 
have the same {\bf signature} if they are indistinguishable in their
ability to combine with other analyses; if so, the parser discards all but the
highest-scoring one.  CKY requires $O(n^3s^2)$ time and $O(n^2s)$
space, where $n$ is the length of the sentence and $s$ is an upper
bound on signatures per substring.

\begin{figure}
\centerline{\psfig{figure=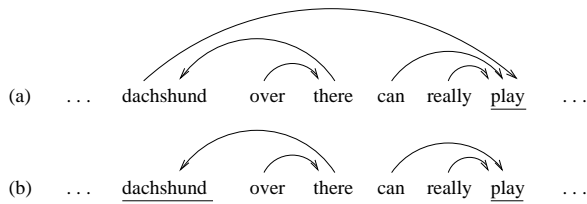,width=3in}}
\caption{\small Spans participating in the correct parse of {\em That dachshund over there
can really play golf!}.  (a) has one parentless endword; its subspan (b) has two.}\label{twospans}
\end{figure}

Let us consider dependency parsing in this framework.  One might guess
that each substring analysis should be a lexical tree---a tagged
headword plus all lexical subtrees dependent upon it.  (See
Figure~\ref{sampleparse}b%
.)  However, if a constituent's probabilistic behavior depends on its
headword---the lexicalist hypothesis---then differently headed
analyses need different signatures.  There are at least $k$ of these
for a substring of length $k$, whence the bound $s = k = \Omega(n)$,
giving a time complexity of $\Omega(n^5)$.
\cite{Collins} uses this $\Omega(n^5)$ algorithm directly (together
with pruning).

We propose an alternative approach that preserves the $O(n^3)$ bound.
Instead of analyzing substrings as lexical trees that will be linked
together into larger lexical trees, the parser will analyze them as
non-constituent {\bf spans} that will be concatenated into larger
spans.  A span consists of $\geq 2$ adjacent words; tags for all
these words except possibly the last; a list of all
dependency links among the words in the span; and perhaps some other
information carried along in the span's signature.  No cycles,
multiple parents, or crossing links are allowed in the span, and each
{\bf internal word} of the span must have a parent in the span.

Two spans are illustrated in Figure~\ref{twospans}.  These diagrams
are typical: a span of a dependency parse may consist of either a
parentless {\bf endword} and some of its descendants on one side
(Figure~\ref{twospans}a), or two parentless endwords, with all the
right descendants of one and all the left descendants of the other
(Figure~\ref{twospans}b).  The intuition is that the internal part of
a span is grammatically inert: except for the endwords {\em dachshund}
and {\em play}, the structure of each span is irrelevant to the span's
ability to combine in future, so spans with different internal
structure can compete to be the best-scoring span with a particular
signature.  

\begin{figure}
\centerline{\psfig{figure=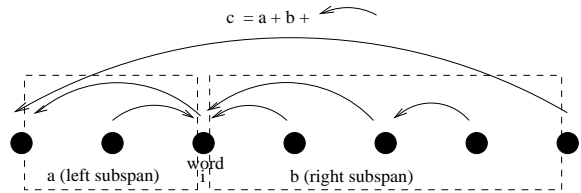,width=3in}}
\caption{\small The assembly of a span $c$ from two smaller spans ($a, b$) and a covering link.  Only $b$ isn't minimal.}\label{assembly}
\end{figure}

If span $a$ ends on the same word $i$ that starts span $b$,
then the parser tries to combine the two spans by 
{\bf covered-concatenation} (Figure~\ref{assembly}).  The two copies
of word $i$ are identified, after which a leftward or rightward {\bf
covering link} is optionally added between the endwords of the new
span.  Any dependency parse can be built up by covered-concatenation.
When the parser covered-concatenates $a$ and $b$, it obtains up
to three new spans (leftward, rightward, and no covering link).  

\begin{figure*}
\begin{eqnarray}
\prod_{k \leq i < \ell} Pr({\sl tword}(i) \mid {\sl tword}(i+1), {\sl tword}(i+2)) 
\cdot \makebox[0.9cm]{$\displaystyle \prod_{k \leq i,j \leq \ell {\rm\ with\ }i,j{\rm\ linked\ }}$} \Pr(\mbox{$i$ has prefs that $j$ satisfies} \mid {\sl tword}(i), {\sl tword}(j)) \label{assumps-times-adverts} \\
\makebox[0.9cm]{$\displaystyle \prod_{k \leq i,j \leq \ell {\rm\ with\ }i,j{\rm\ linked\ }}$} \Pr(L_{ij} \mid {\sl tword}(i), {\sl tword}(j), {\sl tag}({\sl next\mbox{-}closest\mbox{-}kid}(i)))
\cdot \makebox[0.9cm]{$\displaystyle \prod_{k < i < \ell,\ (j < k {\rm\ or\ } \ell < j)}$} \Pr(L_{ij} \mid  {\sl tword}(i), {\sl tword}(j), \cdots) \label{inwordsout-times-edgewordsin} 
\end{eqnarray}
\end{figure*}

The covered-concatenation of $a$ and $b$, forming $c$, is barred unless
it meets certain simple tests:

\noindent $\bullet$ $a$ must be {\bf minimal} (not itself expressible as a concatenation of narrower spans).
   This prevents us from assembling $c$ in multiple ways.

\noindent $\bullet$ Since the overlapping word will be internal to $c$, it must
have a parent in exactly one of $a$ and $b$.

\noindent $\bullet$ $c$ must not be given a covering link if either
the leftmost word of $a$ or the rightmost word of $b$ has a parent.
(Violating this condition leads to either multiple parents or link
cycles.)


Any sufficiently wide span whose left endword has a parent is a
legal parse, rooted at the EOS mark (Figure~\ref{sampleparse}).
Note that a span's signature must specify whether its endwords
have parents.

\section{Bottom-Up Probabilities}

Is this one parser really compatible with all three probability
models?  Yes, but for each model, we must provide a way to keep track
of probabilities as we parse.  Bear in mind that models A, B, and C do
not themselves specify probabilities for all spans; intrinsically they
give only probabilities for sentences.

{\bf Model C.} Define each span's {\bf score} to be the product of all
probabilities of links within the span.  (The link to $i$ from its
$c$th child is associated with the probability $Pr(\ldots)$ in
\refp{genmodelparseprob}.)  When spans $a$ and $b$ are combined and
one more link is added, it is easy to compute the resulting span's
score: ${\rm score}(a) \cdot {\rm score}(b) \cdot Pr({\rm covering\ link})$.%
\footnote{The third factor depends on, e.g., ${\sl kid}(i,c-1)$, which
we recover from the span signature.  Also, matters are complicated
slightly by the probabilities associated with the generation of STOP.
%
%
}

When a span constitutes a parse of the whole input sentence, its score
as just computed proves to be the parse probability, conditional on 
the tree root EOS, under model C.  The highest-probability parse can
therefore be built by dynamic programming, where we build and retain
the highest-scoring span of each signature.

{\bf Model B.}  Taking the Markov process to generate (tag,word) pairs
from right to left, we let \refp{assumps-times-adverts} define the score
of a span from word $k$ to word $\ell$.  The first product encodes
the Markovian probability that the (tag,word)
pairs $k$ through $\ell-1$ are as claimed by the span, conditional on
the appearance of specific (tag,word) pairs at $\ell, \ell+1$.%
\footnote{Different $k$--$\ell$ spans have scores conditioned on
different hypotheses about ${\sl tag}(\ell)$ and ${\sl tag}(\ell+1)$;
their signatures are correspondingly different.  Under model B, a
$k$--$\ell$ span may not combine with an $\ell$--$m$ span whose tags
violate its assumptions about $\ell$ and $\ell+1$.}  Again, scores can
be easily updated when spans combine, and the probability of a 
complete parse $P$, divided by the total probability of all parses
that succeed in satisfying lexical preferences, is just $P$'s score.

{\bf Model A.}  Finally, model A is scored the same as model B, except
for the second factor in \refp{assumps-times-adverts}, which is
replaced by the less obvious expression in
\refp{inwordsout-times-edgewordsin}.  As usual, scores can be
constructed from the bottom up (though ${\sl tword}(j)$ in the second
factor of \refp{inwordsout-times-edgewordsin} is not available to the
algorithm, $j$ being outside the span, so we back off to ${\sl
word}(j)$).

\section{Empirical Comparison}

We have undertaken a careful study to compare these models' success at
generalizing from training data to test data.  Full results on a
moderate corpus of 25,000+ tagged, dependency-annotated {\em Wall
Street Journal} sentences, discussed in \cite{EisnerTR}, were not
complete at press time.  However,
Tables~\ref{resulttags}--\ref{resultpars} show pilot results for a
small set of data drawn from that corpus.
(The full results show substantially better performance, e.g., 93\%
correct tags and 87\% correct parents for model C, but appear
qualitatively similar.)

The pilot experiment was conducted on a subset of 4772 of the
sentences comprising 93,360 words and punctuation marks.  The corpus
was derived by semi-automatic means from the Penn Treebank; only
sentences without conjunction were available (mean length=20, max=68).
A randomly selected set of 400 sentences was set aside for testing all
models; the rest were used to estimate the model parameters.  In the
pilot (unlike the full experiment), the parser was instructed to
``back off'' from all probabilities with denominators $< 10$.  For
this reason, the models were insensitive to most lexical distinctions.

In addition to models A, B, and C, described above, the pilot
experiment evaluated two other models for comparison.  Model C$'$ was
a version of model C that ignored lexical dependencies between parents
and children, considering only dependencies between a parent's tag and
a child's tag.  This model is similar to the model used by stochastic
CFG.  Model X did the same $n$-gram tagging as models A and B ($n=2$
for the preliminary experiment, rather than $n=3$), but did not assign
any links.


\begin{table}
\scriptsize
\begin{tabular}{|l||r|r|r|r|r|r|}\hline
                & A    & B    & C          & C$'$  & X    & Basel.    \\ \hline\hline    
All tokn       & 90.2 & 90.9 &      90.8 & 90.5 & {\bf 91.0} & {\bf 79.8} \\  \hline
{\bf Non-punc} & 88.9 & 89.8 & {\bf 89.6} & 89.3 & {\bf 89.8} & 77.1   \\ \hline 
Nouns           & 90.1 & 89.8 &      90.2  & 90.4 & 90.0 & 86.2   \\ \hline 
Lex verbs   & 74.6 & 75.9 &      73.3  & 75.8 & 73.3 & 67.5   \\  \hline 
\end{tabular}
\caption{Results of preliminary experiments: Percentage of tokens correctly tagged by each model.}\label{resulttags}
\end{table}

\begin{table}
\scriptsize
\begin{tabular}{|l||r|r|r|r|r|}\hline
                & A    & B    & C          & C$'$   & Baseline \\ \hline\hline
All tokens       & 75.9 & 72.8 & {\bf 78.1}&  66.6 & 47.3 \\ \hline
{\bf Non-punc} & 75.0 & 75.4 & {\bf 79.2} & {\bf 68.8} & {\bf 51.1}     \\  \hline 
Nouns           & 75.7 & 71.8 & {\bf 77.2} & 55.9 & 29.8 \\ \hline 
Lexical verbs   & 66.5 & 63.1 & {\bf 71.0} & 46.9 & 21.0  \\ \hline 
\end{tabular}
\caption{Results of preliminary experiments: Percentage of tokens correctly attached to their parents by each model.}\label{resultpars}
\end{table}

Tables~\ref{resulttags}--\ref{resultpars} show the percentage of raw tokens
that were correctly tagged by each model, as well as the proportion
that were correctly attached to their parents.  For tagging,  baseline performance
was measured
by assigning each word in the test set its most frequent tag (if any)
from the training set.  The unusually low baseline performance results
from a combination of a small pilot training set and a mildly extended
tag set.\footnote{We used distinctive tags for auxiliary verbs and for
words being used as noun modifiers (e.g., participles), because they
have very different subcategorization frames.}  We observed that in
the training set, determiners most commonly pointed to the following
word, so as a parsing baseline, we linked every test determiner to the
following word; likewise, we linked every test preposition to the preceding
word, and so on.

The patterns in the preliminary data are striking, with verbs showing
up as an area of difficulty, and with some models clearly faring
better than other.  The simplest and fastest model, the recursive
generation model C, did easily the best job of capturing the
dependency structure (Table~\ref{resultpars}).  It misattached the
fewest words, both overall and in each category.  This suggests that
subcategorization preferences---the only factor considered by model
C---play a substantial role in the structure of Treebank sentences.
(Indeed, the errors in model B, which performed worst across the
board, were very frequently arity errors, where the desire of a child
to attach to a particular parent overcame the reluctance of the parent
to accept more children.)

A good deal of the parsing success of model C seems to have arisen
from its knowledge of individual words, as we expected.  This is shown
by the vastly inferior performance of the control, model C$'$.  On the
other hand, both C and C' were competitive with the other models at
tagging.  This shows that a tag can be predicted about as well from
the tags of its putative parent and sibling as it can from the tags of
string-adjacent words, even when there is considerable error in
determining the parent and sibling.

\section{Conclusions}

Bare-bones dependency grammar---which requires no link labels, no
grammar, and no fuss to understand---is a clean testbed for studying
the lexical affinities of words.  We believe that this is an important
line of investigative research, one that is likely to produce both
useful parsing tools and significant insights about language modeling.

As a first step in the study of lexical affinity, we asked whether
there was a ``natural'' way to stochasticize such a simple formalism
as dependency.  In fact, we have now exhibited three promising types
of model for this simple problem.  Further, we have developed a novel
parsing algorithm to compare these hypotheses, with results that so
far favor the speaker-oriented model C, even in written, edited {\em
Wall Street Journal} text.  To our knowledge, the relative merits of
speaker-oriented versus hearer-oriented probabilistic syntax models
have not been investigated before.


\begin{thebibliography}{}
\small
\bibitem[\protect\citename{Black \bgroup et al.\egroup}1992]{Black+al}
  Ezra Black, Fred Jelinek, et al.
\newblock 1992.
\newblock Towards history-based grammars: using richer models for probabilistic parsing.
\newblock In {\em Fifth DARPA Workshop on Speech and Natural Language}, Arden 
  Conference Center, Harriman, New York, February.
\newblock cmp-lg/9405007.

\bibitem[\protect\citename{Church}1988]{Church:trigram}
  Kenneth~W. Church.
\newblock 1988.
\newblock A stochastic parts program and noun phrase parser for unrestricted
  text.
\newblock In {\em Proc. of the 2nd Conf.\ on Applied Natural
  Language Processing}, 136--148, Austin, TX.  Association for
  Computational Linguistics, Morristown, NJ.

\bibitem[\protect\citename{Collins}1996]{Collins} Michael~J. Collins.
  \newblock 1996.  \newblock A new statistical parser based on bigram
  lexical dependencies.  \newblock {\em Proc.\ of the 34th
  ACL}, Santa Cruz, July.
\newblock cmp-lg/9605012.

\bibitem[\protect\citename{Eisner}1996]{EisnerTR} 
  Jason Eisner.
\newblock 1996.  
\newblock An empirical comparison of probability models for
dependency grammar.  Technical report IRCS-96-11, University of Pennsylvania.  
\newblock cmp-lg/9706004.

\bibitem[\protect{Jelinek}1985]{Jelinek:trigram}
  Fred Jelinek.
\newblock 1985. 
\newblock Markov source modeling of text generation.
\newblock In J. Skwirzinski, editor, {\em Impact of Processing Techniques on Communication},
  Dordrecht.

\bibitem[\protect{Jelinek \bgroup et al.\egroup}1992]{Jelinek+al}
  Fred Jelinek, John~D. Lafferty, and Robert L.~Mercer.
\newblock 1992.
\newblock Basic methods of probabilistic context-free grammars.
\newblock In {\em Speech Recognition and Understanding: Recent
  Advances, Trends, and Applications}.


\bibitem[\protect{Kupiec}1992]{Kupiec:trigram}
J. Kupiec.  
\newblock 1992.
\newblock Robust part-of-speech tagging using a hidden Markov model.
\newblock {\em Computer Speech and Language}, 6.

\bibitem[\protect\citename{Lafferty \bgroup et al.\egroup}1992]{Lafferty+al}
John Lafferty, Daniel Sleator, and Davy Temperley.
\newblock 1992.
\newblock Grammatical trigrams: A probabilistic model of link grammar
\newblock In {\em Proc. of the AAAI Conf. on
Probabilistic Approaches to Natural Language}, Oct.

\bibitem[\protect\citename{Magerman}1995]{Magerman}
David Magerman.
\newblock 1995.
\newblock Statistical decision-tree models for parsing.
\newblock In {\em Proceedings of the 33rd ACL}, 
Boston, MA.  
\newblock cmp-lg/9504030.

\bibitem[\protect{Mel'\v{c}uk}1988]{Melcuk}
Igor~A. Mel'\v{c}uk.
\newblock 1988.
\newblock {\em Dependency Syntax: Theory and Practice.}
\newblock State University  of New York Press.

\bibitem[\protect{Merialdo}1990]{Merialdo:trigram}
B. Merialdo.
\newblock 1990.
\newblock Tagging text with a probabilistic model.
\newblock In {\em Proceedings of the IBM Natural Language ITL,} Paris, France, pp. 161-172.

\bibitem[\protect\citename{Schabes}1992]{Schabes}
Yves Schabes.
\newblock 1992.
\newblock Stochastic lexicalized tree-adjoining grammars.
\newblock In {\em Proceedings of COLING-92}, Nantes, France, July.

\bibitem[\protect\citename{Sleator and Temperley}1991]{Sleator+Temperley}
Daniel Sleator and Davy Temperley.
\newblock 1991.
\newblock Parsing English with a Link Grammar.
\newblock Tech. rpt. CMU-CS-91-196.
\newblock Carnegie Mellon Univ.
\newblock cmp-lg/9508004.

\end{thebibliography}
\end{document}